\documentclass{article}

\usepackage{arxiv}

\usepackage[utf8]{inputenc} % allow utf-8 input
\usepackage[T1]{fontenc}    % use 8-bit T1 fonts
\usepackage{hyperref}       % hyperlinks
\usepackage{url}            % simple URL typesetting
\usepackage{booktabs}       % professional-quality tables
\usepackage{amsfonts}       % blackboard math symbols
\usepackage{nicefrac}       % compact symbols for 1/2, etc.
\usepackage{microtype}      % microtypography
\usepackage{lipsum}
\usepackage{siunitx}
\usepackage{pgf}
\usepackage{tikz}
\usetikzlibrary{arrows,automata}
\usepackage{amsmath}
\usepackage{mathtools}

\newcommand{\T }{^{\text{T}}}

\DeclarePairedDelimiter{\normtwo}{\lVert}{\rVert}

\let\vec\mathbf
\usepackage{algorithm}
\usepackage{algorithmic}
\usepackage{amssymb}
\usepackage[misc]{ifsym}

\title{Graph-based calibration transfer}

\author{
  Ramin Nikzad-Langerodi, PhD\\
  Software Competence Center Hagenberg\\
  Softwarepark 21\\
  Hagenberg, 4232, Austria\\
  \texttt{ramin.nikzad-langerodi@scch.com} \\
   \And
  Florian Sobieczky, PhD\\
  Software Competence Center Hagenberg\\
  Softwarepark 21\\
  Hagenberg, 4232, Austria\\
  \texttt{florian.sobieczky@scch.com}
}

\begin{document}
\maketitle

\begin{abstract}
The problem of transferring calibrations from a primary to a secondary instrument, i.e. calibration transfer (CT), has been a matter of considerable research in chemometrics over the past decades. Current state-of-the-art (SoA) methods like (piecewise) direct standardization perform well when suitable transfer standards are available. However, stable calibration standards that share similar (spectral) features with the calibration samples are not always available. Towards enabling CT with arbitrary calibration standards, we propose a novel CT technique that employs manifold regularization of the partial least squares (PLS) objective. In particular, our method enforces that calibration standards, measured on primary and secondary instruments, have (nearly) invariant projections in the latent variable space of the primary calibration model. Thereby, our approach implicitly removes inter-device variation in the predictive directions of $\vec X$ which is in contrast to most state-of-the-art techniques that employ explicit pre-processing of the input data. We test our approach on the well-known corn benchmark data set employing the NBS glass standard spectra for instrument standardization and compare the results with current SoA methods\footnote{The research reported in this work has been partly funded by BMVIT, BMDW, and the Province of Upper Austria in the frame of the COMET Program managed by the FFG. The COMET Centre CHASE is funded within the framework of COMET - Competence Centers for Excellent Technologies by BMVIT, BMDW, the Federal Provinces of Upper Austria and Vienna. The COMET program is run by FFG.}.  
\end{abstract}

% keywords can be removed

\section{Introduction}
\label{sec:introduction}
Calibration transfer (CT), sometimes referred to as instrument standardization in chemometrics, is the process of transferring a calibration model from one instrument to another \cite{Niranalysis,FEUDALE2002181,Workman:18}. Ideally, CT preserves the accuracy and precision of a calibration model developed on a primary instrument, i.e. providing statistically identical analysis of the same samples measured on the secondary instrument. Historically, CT has been addressed by i) model updating, or ii) measuring a set of so-called calibration standards on both instruments in order to derive a correction for the difference in the instrumental response. The former can be considered more generic and copes with any type of change related to the measurement condition such as environmental influences, matrix effects or instrumental changes. Slope and bias correction \cite{doi:10.1021/ac9510595}, calibration set augmentation \cite{NikzadLughoferCernudaReischerKantnerPawliczekBrandstetter18}, model updating via Tihkonov regularization \cite{doi:10.1366/000370209788701206} or domain-invariant modelling \cite{doi:10.1002/cem.2818,doi:10.1021/acs.analchem.8b00498,8999284} all belong to this category and have been applied with success to CT problems. However, these methods usually require a considerable amount of (additional) samples (and reference measurements) and deciding between maintenance and re-calibration is not always straightforward. Calibration transfer by means of calibration standards, on the other hand, solely requires a small set of samples that can be measured on both devices and does not require any additional reference values. In this second category, Direct- (DS) and piecewise direct standardization (PDS) can be considered the gold standard \cite{Wang1991}. Both operate by learning a multivariate transformation (mapping) such that the instrumental response of the secondary instrument matches with the one of the primary instrument. Several alternative techniques have been recently proposed such as: Generalized least squares weighting (GLSW) which down-weighs directions in the predictor matrix that are associated with large between-device differences \cite{wise2001calibration,doi:10.1002/cem.780}; orthogonal signal correction (OSC) and related techniques based on orthogonalization of the regression vector to differences between devices \cite{SJOBLOM1998229, IGNE200957}; spectral subspace transform (SST) which reconstructs the secondary instruments' response in the space spanned by the primary instrument's principle components \cite{DU201164}; spectral regression (SR) and PLSCT, which perform DS between (non-linear) embeddings of the calibration standards \cite{PENG20111315,Zhao2019} and data integration methods such as Joint and Unique Multiblock Analysis (JUMBA) \cite{Skotare2019} or Joint-Y partial least squares (PLS) \cite{doi:10.1002/cem.2874} that aim at deriving models based on latent variables (LVs) that are common to primary and secondary instrumental responses. 

Looking at the vast majority of publications that have addressed CT over the past decades, two general observations can be made: Most techniques seem to work well when the standardization set is a (carefully selected) subset of the calibration set, i.e. if samples and standards share similar spectral features. In addition, most CT methods proposed so far perform explicit corrections of the instrumental response rather then modelling the inter-device variation implicitly. 

Towards CT with arbitrary calibration standards, we thus propose a method based on a novel, regularized PLS variant that implicitly corrects for inter-device variation in the directions of the input matrix that are related to the target property. In particular, our approach aims at preserving the special graph in which only the matched calibration standards are connected by a vertex while finding an embedding of the (primary) calibration samples that is informative w.r.t. the target property. Along this line our approach can be considered a special case of previous work on locality preserving projection \cite{10.5555/2981345.2981365} in the context of PLS modelling \cite{doi:10.1021/acs.iecr.5b02559,8859330}.     

\section{Theory}
\label{sec:Theory}

\subsection{Notation}
We follow standard notation in chemometrics, where upper and lower case boldface symbols denote matrices and vectors, respectively, and scalars are denoted by non-boldface symbols. Unless otherwise stated, vectors are column vectors and $\T$ and $^{-1}$ denote the transpose and inverse operation, respectively. By $\vec I$ and $\boldsymbol 0$ we denote the identity and Null matrix (with appropriate dimensions), respectively, and by $\otimes$ the tensor (Kronecker) product between two matrices. Without loss of generality we assume that the response $y$ is univariate. The matrices $\vec A$, $\vec D$ and $\vec L$ are used to denote the adjacency, degree and Laplace matrix of a simple, undirected graph and $\normtwo{\cdot}_\text{F}$ denotes the Frobenius norm. Comma and semicolon notations are used to denote horizontal and vertical stacking of scalars and matrices, e.g. $[\vec x, y]$ and $[\vec x;\vec y]$.    

\subsection{Problem formulation}
\label{sec: Problem formulation}
We describe our method by taking the situation where there are $N$ labeled calibration samples from the primary instrument, i.e. $\{(\vec x_i, y_i)\in\mathbb{R}^{d}\times\mathbb{R}|\;i=1,\dots,N\}$, and a set of $K$ calibration standard samples $\{(\vec x^{(p)}_i, \vec x^{(s)}_i)\in \mathbb{R}^{d}\times\mathbb{R}^{d}|\;i=1,\dots,K\}$,  measured on the primary and secondary instrument. It is assumed that 
the same $d$ variables are measured on both primary and secondary instrument, i.e. that the wavelength axis is stable across the instruments. The goal of calibration transfer is to derive a model $h:\mathbb{R}^d\rightarrow\mathbb{R}$ from the primary calibration and the calibration standard samples that has a small error when applied to samples measured on the secondary instrument.

\subsection{Graph-based calibration transfer}
In order to minimize inter-device variation when building the primary calibration model, we proceed as following: We let $\vec X_p = [\vec x_1^{(p)},\vec x_2^{(p)},\dots,\vec x_K^{(p)}]\T$ and $\vec X_s = [\vec x_1^{(s)},\vec x_2^{(s)},\dots,\vec x_K^{(s)}]\T$ be matrices holding the matched calibration standards measured on the primary and the secondary instrument, respectively. We further let $\vec K = [\vec X_p;\vec X_s] = [\vec k_1,\dots,\vec k_{2K}]^T$, $\vec X$ and $\vec y$ hold the calibration samples from the primary instrument, and
\begin{equation}
\label{eqn:regularizer}
\begin{aligned}
J  & =  \frac12\sum_{i=1}^{2K}\sum_{j=1}^{2K} (\vec k_i\T\vec w - \vec k_j\T\vec w)^2\;\vec A_{i,j}\\
& =  \vec w\T \left[ \sum_{i=1}^{2K} \vec k_i\vec D_{i,i}\vec k_i\T - \sum_{i=1}^{2K}\sum_{j=1}^{2K} \vec k_i\vec A_{i,j}\vec k_j\T \right] \vec w\\
& = \vec w\T\vec K\T(\vec D - \vec A)\vec K\vec w\\
& = \vec w\T \vec K\T \vec L \vec K\vec w,
\end{aligned}
\end{equation}
with $\vec D = \vec I$, $\vec A = \begin{pmatrix}
0 & 1\\
1 & 0 \end{pmatrix}\otimes \vec I$, $\vec L = \begin{pmatrix}
1 & -1\\
-1 & 1 \end{pmatrix}\otimes \vec I$ and $\vec I$ denoting the $K\times K$ identity matrix. We then compute
\begin{align}
\label{eq:close_form}
\vec w^{*} := \left[\frac{\vec y\T\vec X}{\vec y\T\vec y}\left(\vec I + \frac{\gamma}{\vec y\T\vec y}\boldsymbol{\Gamma}\right)^{-1}\right]\T
\end{align}
with $\boldsymbol \Gamma = \vec K\T\vec L\vec K$, which is the minimizer of (see Appendix for derivation)
\begin{align}
\label{eq:objective_function}
\min_{\vec w}\,\normtwo{\vec X - \vec y \vec w\T}^2_\text{F} + \gamma J,
\end{align}
where $\gamma$ is a regularization parameter that controls the trade-off between attaining a subspace that is predictive w.r.t. the response $\vec y$ of the (primary) calibration samples and minimizing the Euclidean distance between the matched calibration transfer samples in the latent variable space. The first term in Eq.\eqref{eq:objective_function} maximizes the covariance between $\vec X$ and $\vec y$ and is thus equivalent to the partial least squares objective. The second term corresponds to a so-called manifold regularization term and biases the PLS weight vector $\vec w$ such that the inter-device variation becomes smaller. After computing $\vec w^{*}$ we proceed in analogy with the well-known NIPALS algorithm, i.e. we perform following iterations:
\begin{enumerate}
	\item Mean centering $\vec X = \vec X_0(\vec I -\frac{1}{N}\boldsymbol{1}\boldsymbol{1}\T) ; \vec y = \vec y_0 - \bar{\vec y}$, where $ \vec X_0$ and $\vec y_0$ refer to the raw data,
	\item Stack (deflated) calibration standard matrices $\vec K = [\vec X_p;\vec X_s]$
	\item Compute regularization term $\boldsymbol{\Gamma} = \vec K\T\vec L\vec K$  
	\item Compute weight vector $\vec w^{*} = \left[\frac{\vec y\T\vec X}{\vec y\T\vec y}\left(\vec I + \frac{\gamma}{\vec y\T\vec y}\boldsymbol{\Gamma}\right)^{-1}\right]\T$
	\item Normalize $\vec w^{*} = \frac{\vec w^{*}}{\normtwo{\vec w^{*}}}$
	\item Compute of scores $\vec t = \vec X\vec w^* ; \vec t_p = \vec X_p\vec w^* ; \vec t_s = \vec X_s\vec w^*$
	\item Regression $c = \frac{\vec t\T \vec y}{\vec t\T\vec t}$
	\item Compute loadings $\vec p = \frac{\vec X\T\vec t}{\vec t\T\vec t} ; \vec p_p = \frac{\vec X_p\T\vec t_p}{\vec t_p\T\vec t_p} ; \vec p_s = \frac{\vec X_s\T\vec t_s}{\vec t_s\T\vec t_s}$
	\item Deflation: Transform $\vec X$ into $ \vec X - \vec t\vec p\T$, $\vec X_p$ into $\vec X_p - \vec t_p\vec p_p\T$, and $\vec X_s$ into $  \vec X_s - \vec t_s\vec p_s\T$
	\item Aggregation: Transform $\vec W$ into $\vec [\vec W, \vec w^*] $, $\vec P$ into  $ \vec [\vec P, \vec p] $, and $\vec c$ into $\vec [\vec c; c]$
	\item Return to 2 until $A$ LVs have been computed
	\item Compute regression coefficients $\vec b = \vec W(\vec P\T\vec W)^{-1}\vec c$
\end{enumerate}

\begin{figure}[h!]
	\centering
	\begin{tikzpicture}[-,shorten >=1pt,auto,node distance=1.5cm,scale=0.50]
	\tikzstyle{every state}=[text=black]
	
	\node[state]         (A) [fill=red]       {$1$};
	\node[state]         (B) [above right of=A, fill=blue] {$1^\prime$};
	\node[state]         (D) [below right of=A, fill=red] {$2$};
	\node[state]         (C) [below right of=B, fill=blue] {$2^\prime$};
	
	\node[state]         (E) [right of=C, draw=none] {};
	\node[state]         (F) [below of=D, draw=none] {};
	
	\node[state]         (G) [left of=A, draw=none] {};
	
	\path (A) edge              node {}(B)
	(C) edge 			  node {}(D)
	(E) edge [<-] node {$\vec w_1$}(F)
	(F) edge [->] node {$\vec w_2$}(G);
	\end{tikzpicture}
	\caption{Illustrative example of the idea of graph-based calibration transfer (GCT) with two calibration transfer samples measured on the primary (red) and secondary (blue) instrument. Only matched samples (i.e. the same sample measured on both instruments) have a vertex. Projection on direction $\vec w_1$ and $\vec w_2$ results in large and small inter-device difference in the LV space.}
	\label{fig:expl}
\end{figure}
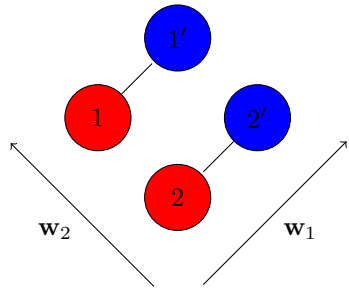

\subsection{Discussion}
As an illustrative example consider the situation of having exactly two calibration standard samples (Figure \ref{fig:expl}). For the adjacency matrix $\vec A$ in Eq.\eqref{eqn:regularizer} we then have
\begin{align*}
\label{eqn:adjacency}
\vec A &= \begin{bmatrix}
\boldsymbol 0 & \vec I\\
\vec I & \boldsymbol 0
\end{bmatrix}
= \begin{bmatrix}
0 & 0 & 1 & 0\\
0 & 0 & 0 & 1\\
1 & 0 & 0 & 0\\
0 & 1 & 0 & 0
\end{bmatrix},
\end{align*}  
i.e. only matched standard samples have a vertex in the corresponding graph defined over $\vec K$. Consequently, the regularization term in Eq. \eqref{eq:objective_function} is large if the Euclidean distances btw. the projections of the matched standard samples are large. Thus, increasing the magnitude of the parameter $\gamma$ reduces the difference between primary and secondary instruments in the $y$-predictive direction of $\vec X$. Since the Laplacian matrix $\vec L = \vec 
D - \vec A$ is always positive semi-definite, the regularizer $\vec w\T\boldsymbol\Gamma \vec w$ is convex and preserves the convexity of the objective function \cite{8859330}. Thus, the unique solution of Eq.\eqref{eq:objective_function} is attained as the root of the derivative and has the closed-form solution of Eq. \eqref{eq:close_form}.

\section{Experimental section}
\subsection{Data set}
 All experiments were carried out using the Cargill corn benchmark data set obtained from Eigenvector Research Inc.\footnote{http://www.eigenvector.com/data/Corn/ (accessed April 11, 2018)}. The data set comprises NIR spectra from a set of 80 samples of corn measured on 3 different spectrometers at 700 spectral channels in the wavelength range 1100-2498 nm at 2 nm intervals. In the present work, we considered calibration transfer between instruments m5 and mp6 for prediction
 of moisture, oil, protein and starch. The spectra of the first 3 NBS glass standards included in the data set were used as calibration standards in all experiments.
 
\subsection{Evaluation}
All experiments were conducted using in-house code implemented in Python programming language. Calibration (60) and validation (20) sets were derived by means of the Kennard-stone algorithm \cite{doi:10.1080/00401706.1969.10490666} from the primary instrument samples. The spectra of the first 3 NBS glass standards measured on the primary and secondary instrument were used together with the calibration set of the primary instrument in order to fit calibration models that were subsequently validated using the validation samples from the secondary instrument. The calibration set was mean centered and the validation set re-centered to the mean of the calibration set. Unless explicitly stated, spectra of the calibration standards were not mean centered. Comparison of our approach with direct standardization (DS)  was carried out by correcting the calibration samples (from the primary instruments) s.t.
\begin{align}
%\label{eqn:direct_standardization}
\tilde{\vec{X}}_{\text{cal}} = \vec{X}_{\text{cal}}\vec F
\end{align}
before model building with $\vec F = \vec X_p^\dagger\vec X_s$ and $\vec X_p^\dagger$ is the pseudo-inverse of the calibration standard spectra from the primary instrument. For CT with corn samples as calibration standards, 10 samples were choosen from the calibration set by means of the Kennard-Stone algorithm.

\section{Results}
\begin{figure}[ht!]
	\centering
	\includegraphics[width=0.99\linewidth]{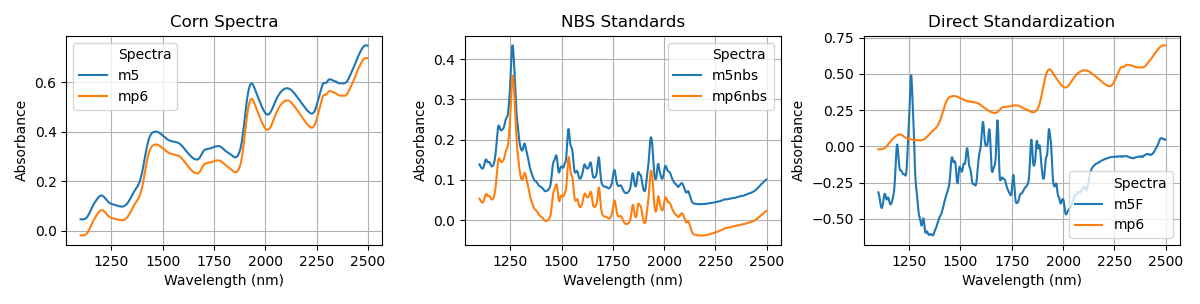}
	\caption{Direct standardization. Left: Calibration  and validation sets corresponding to instruments m5 (blue) and mp6 (orange), respectively. Middle: Calibration standards (NBS glass samples) measured on m5 and mp6. Right: m5 spectra after application of direct standardization (m5F).}
	\label{fig:figure1spectra}
\end{figure}
In most works that have been published on calibration transfer so far, a subset of the calibration samples constitutes the transfer standards. However, this requires that the samples are sufficiently stable, since physical or chemical degradation between measurement on the primary and secondary instrument would lead to deterioration of CT efficiency. For a large variety of sample types, e.g. agricultural or related (food) products, stability is not given. Ideally, CT should therefore be undertaken with stable materials that can have different spectral features compared to the calibration samples. However, standard CT techniques usually break down in such scenarios. As an example, Figure \ref{fig:figure1spectra} shows application of direct standardization (DS) of the corn spectra from instrument m5 using the NBS glass standards m5nbs and mp6nbs. Obviously, the difference in the spectral response on the corn samples can not be corrected for by the glass samples, since the spectral features of the latter are introduced in the standardization. Thus a calibration based on the standardized m5 spectra (i.e. m5F) will not generalize well to samples analyzed on instrument mp6. 

\subsection{Proof of principle}
\begin{figure}[h!]
	\centering
	\includegraphics[width=0.99\linewidth]{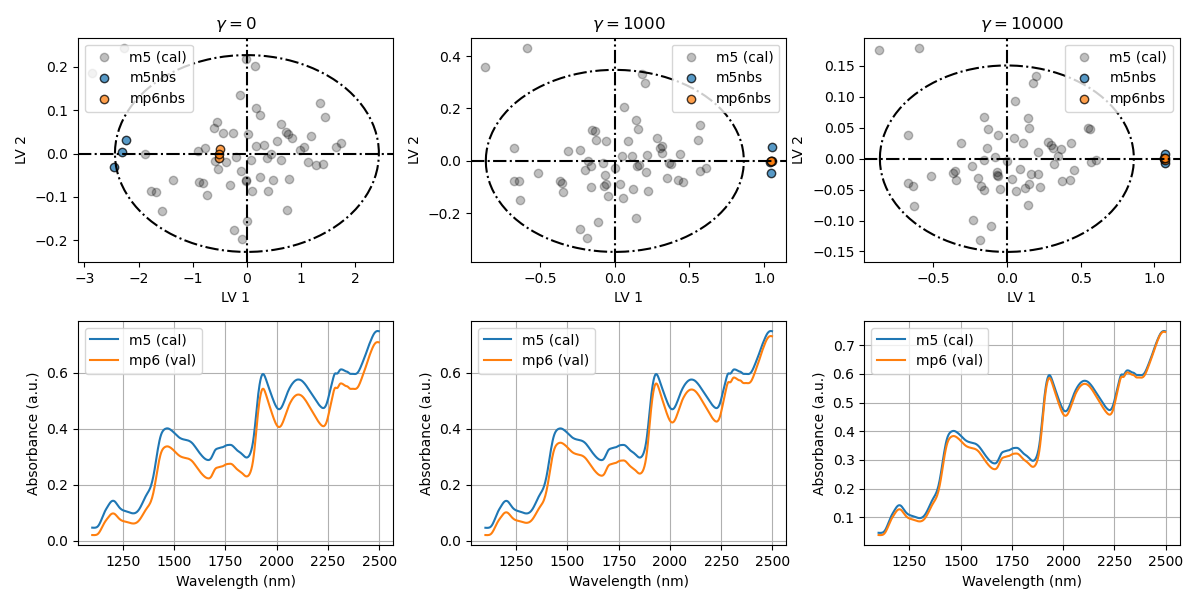}
	\caption{Graph-based calibration transfer. Top panel: Projections of m5 calibration samples (grey), m5nbs and mp6nbs calibration standards with increasing amount of regularization. Bottom panel: Corresponding reconstruction of the mean spectra of m5 calibration and mp6 validation samples from a 2 LV model for the prediction of moisture established using the m5 calibration samples.}
	\label{fig:figure2gct}
\end{figure}

\begin{figure}[h!]
	\centering
	\includegraphics[width=0.99\linewidth]{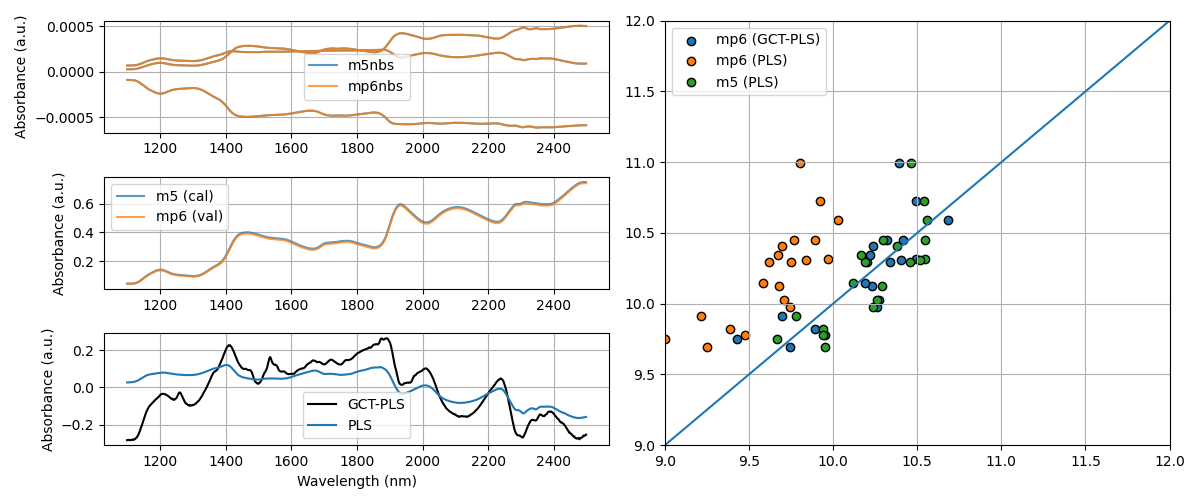}
	\caption{Prediction of moisture. Top left: Reconstructed calibration standard spectra. Note that primary and secondary calibration standard spectra have been (locally) mean centered for better visibility of the reconstructed spectra. Middle left: Reconstructed (mean) corn spectra of the calibration set (m5) and the projected validation samples (mp6). Bottom left: regression coefficients of a 2LV GCT-PLS ($\gamma = 10^6$) and corresponding PLS model. Right: Measured vs. predicted moisture contents. Predictions of the GCT-PLS and PLS models on the mp6 validation samples (blue and orange) and predictions of the latter applied to m5 validation samples (green) are shown.}
	\label{fig:figure4}
\end{figure}

In order to avoid that the spectral features of the calibration standards are carried over to the spectra of the calibration samples, our approach standardizes the instrumental response in the $y$-predictive LV space of the primary calibration model. However, rather than employing DS between the CT samples in the LV space of the primary calibration model as proposed by Zhao and co-workers \cite{Zhao2019}, our approach aims at implicit standardization by jointly mapping the (primary) calibration samples and calibration standards measured on the primary and secondary instrument such that i) the latent variables are predictive w.r.t. the response and ii) the distance between the projections of matched calibration samples is small. Figure \ref{fig:figure2gct} shows the projections of the NBS glass standards measured on m5 and mp6 together with the calibration set from m5 on the first 2 LVs for increasing amount of regularization. Without regularization, the LV space corresponds to the ordinary PLS subspace in which m5nbs and mp6nbs samples differ mostly along the first LV. With increasing $\gamma$, the difference between the calibration standards becomes smaller. Notably, the reconstructed spectra of the calibration samples (measured on m5) and validation samples (measured on mp6) from the corresponding LVs become increasingly similar with increasing regularization (Figure \ref{fig:figure2gct}, bottom panel) demonstrating successful standardization of the corn spectra using the NBS glass standards by the here proposed technique.  
Figure \ref{fig:figure4} shows the same 2 LV model for the prediction of moisture at $\gamma = 10^6$. Both, the reconstructed calibration standard spectra (top left) as well as the reconstructed spectra from the calibration (m5) and the projected validation samples (mp6) are well aligned and the predictions on the mp6 validation samples thus in good accordance with the respective reference measurements (right plot). The regression vector of the GCT-PLS model displays mostly the same features when compared to the unregularized model but has overall higher variance which can be attributed to the alignment of the matched calibration standards in the LV space. Comparing the accuracy on the validation samples, we have RMSEPs of 0.21, 0.61 and 0.20 for the GCT-PLS, PLS (applied to the mp6 validation samples) and PLS model (applied to the m5 validation samples), respectively indicating that the transfer model exhibits similar accuracy to the primary (baseline) model applied to the primary validation samples.

\subsection{Number of LVs}
\begin{figure}[]
	\centering
	\includegraphics[width=0.99\linewidth]{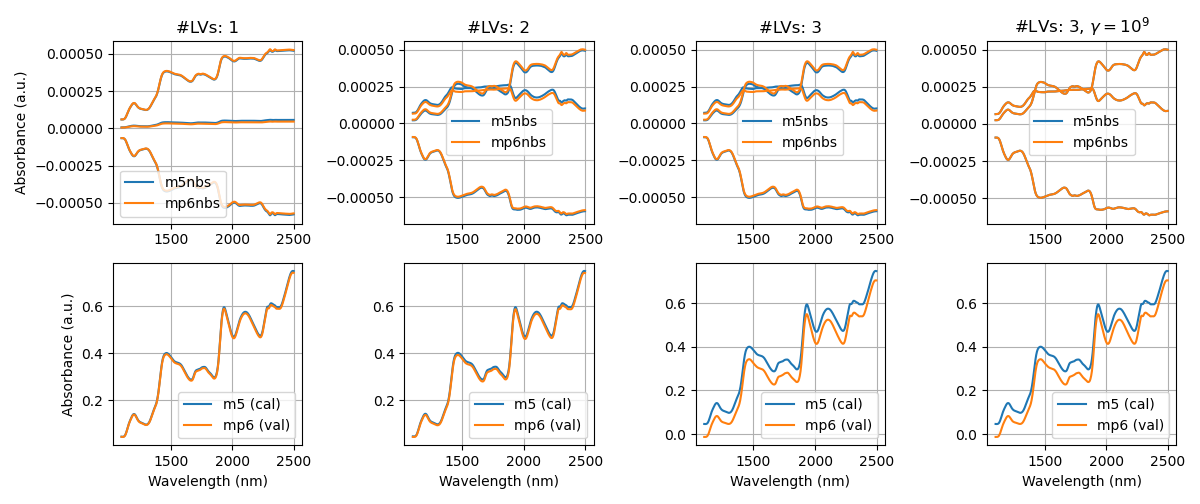}
	\caption{Effect of the number of LVs in GCT. Reconstruced spectra of the calibration standards (top panel) and corn samples (bottom panel) at increasing number of LVs when modeling moisture. The number of LVs is indicated column-wise. For the first 3 columns $\gamma=10^6$.}
	\label{fig:figure5}
\end{figure}

\begin{figure}[]
	\centering
	\includegraphics[width=0.75\linewidth]{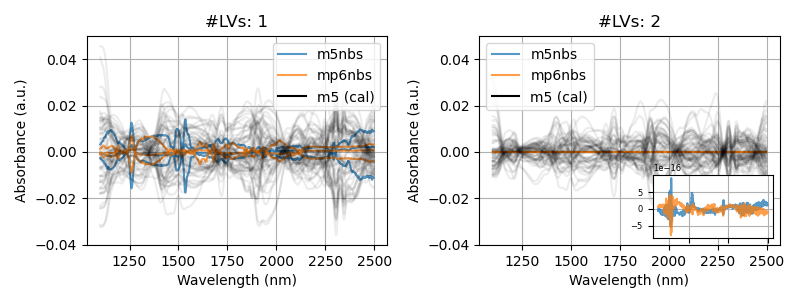}
	\caption{Transfer standard residuals. The residuals of m5nbs and mp6nbs calibration standards and m5 calibration samples are shown after fitting the first (left) and second (right) LV using GCT-PLS ($\gamma = 10^6$) for the prediction of moisture. The calibration standards' residuals are magnified in the inset of the right plot for better visibility.}
	\label{fig:figure6}
\end{figure}

Whether calibration transfer given one type of samples and another type of calibration standards will succeed in practice will largely depend on the analytical accuracy that is aimed at and in turn the number of LVs required to attain that accuracy. Figure \ref{fig:figure5} shows the reconstructed spectra of the NBS calibration standards as well as the m5 calibration and mp6 validation samples when increasing the number of LVs included in the model. For one and two LVs, setting $\gamma = 10^6$ aligns either the (reconstructed) transfer standards' as well as the corn samples' spectra. In contrast, alignment of the standards does not translate into proper standardization of the corn spectra for a 3 LV model - even at stronger regularization (rightmost plots). This finding indicates that beyond 2 LVs there is no structure left in the residuals of the calibration standards' spectra that encodes differences in the instrumental response of primary and secondary instruments. 
Figure \ref{fig:figure6} shows the residuals of the NBS standard samples for a GCT-PLS model with a single and two LVs for the prediction of moisture from the m5 spectra. Obviously, there is no further structure left in the residuals of the NBS standards' spectra after the second round of deflation to correct for the differences between the primary and secondary instrument's response. This result indicates that the calibration standards' residuals represent a useful diagnostic to estimate the number of LVs up to which instrument standardization is feasible given the calibration set (from the primary instrument) and the calibration standards' spectra.

\subsection{Calibration transfer via NBS standards and corn samples}
\begin{figure}[]
	\centering
	\includegraphics[width=0.99\linewidth]{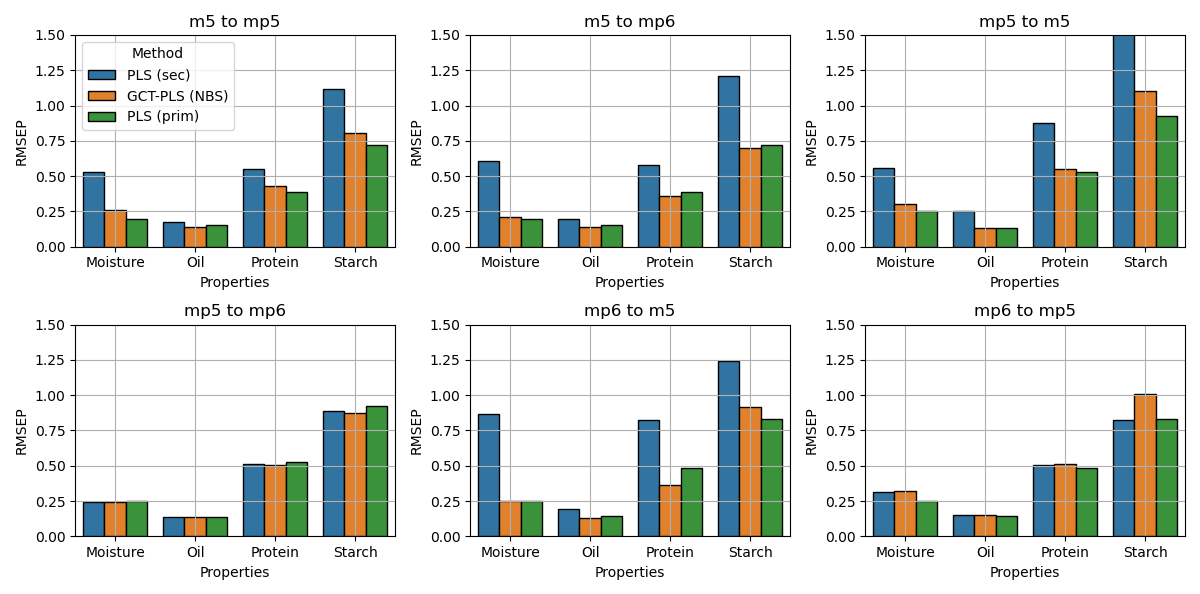}
	\caption{Calibration transfer with NBS standards. Root mean squared errors of prediction (RMSEP) on the validation samples measured on the secondary instrument are shown for all CT scenarios. The accuracy of 2 LV GCT-PLS models with $\gamma = 10^6$ (orange) are compared with standard PLS with 2 LVs (blue). The green bars indicate the accuracy of the primary model evaluated on the primary validation samples.}
	\label{fig:figure8}
\end{figure}

\begin{figure}[]
	\centering
	\includegraphics[width=0.99\linewidth]{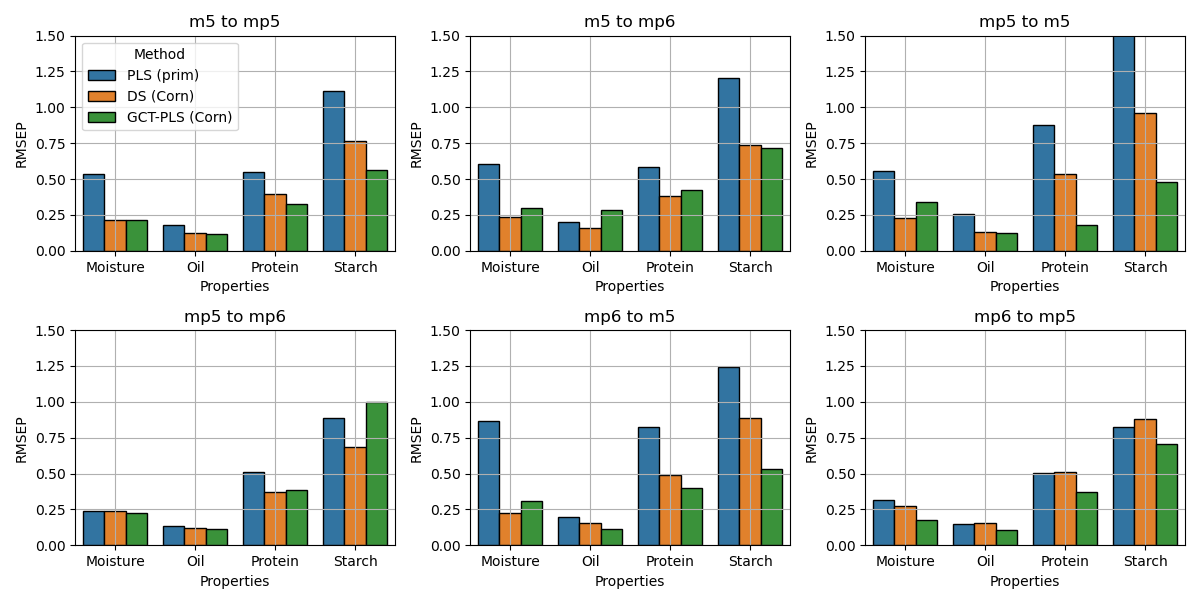}
	\caption{Calibration transfer with corn standards. Root mean squared errors of prediction (RMSEP) on the validation samples measured on the secondary instrument are shown for all CT scenarios. The accuracy of 2 LV GCT-PLS models (green) are compared with standard PLS with 2 LVs (blue) and direct standardization (DS, orange) when using 10 corn samples picked from the calibration set by means of the Kennard-Stone algorithm.}
	\label{fig:figure9}
\end{figure}

In Figure \ref{fig:figure8} we compare the RMSEP of GCT-PLS (using the NBS glass standards as CT samples) on the validation data from the secondary instrument with standard PLS (2 LVs). Except for CT between instruments mp5 and mp6, where the difference in the instrumental response is in general small, the accuracy of GCT-PLS models is significantly higher compared to the baseline PLS models and comparable to the accuracy when the primary PLS model is applied to primary validation samples. These findings demonstrate successful application of GCT-PLS for CT using the NBS glass standards for up to 2 LVs. Finally, we also investigated application of GCT-PLS when corn samples instead of glass samples are used as the calibration standards (Figure \ref{fig:figure9}). Overall we found similar accuracy of GCT-PLS and direct standardization (DS) in most CT scenarios with GCT-PLS outperforming DS when modelling protein and starch contents (except for CT from mp5 to mp6).  

\section{Conclusion}
We have here introduced a graph-based approach to address the calibration transfer problem in analytical chemistry which, unlike most other SoA methods, implicitly standardizes the instrumental responses of primary and secondary instruments while modelling the property of interest. The main benefit of our approach lies in the fact that calibration standards and calibration samples must not share the same spectral features, since the instrumental differences are implicitly modelled in the space of the calibration samples. We have further shown experimentally that the residuals of the calibration standard spectra in the GCT model indicate how many LVs the chosen calibration standards permit to transfer from one instrument to the other, which is an important quantity when assessing the feasibility of CT under given requirements regarding accuracy.

\section*{acknowledgments}
The research reported in this work has been partly funded by BMVIT, BMDW, and the Province of Upper Austria in the frame of the COMET Program managed by the FFG. The COMET Centre CHASE is funded within the framework of COMET - Competence Centers for Excellent Technologies by BMVIT, BMDW, the Federal Provinces of Upper Austria and Vienna. The COMET program is run by FFG.

\bibliographystyle{unsrt}  
\bibliography{references}  %%% Remove comment to use the external .bib file (using bibtex).


\noindent {\bf Appendix}\\

\noindent The calculation of the optimal weights vector $w^*$ is easily obtained from the necessary condition 
\begin{eqnarray}
\nabla_w\left(\normtwo{\vec X \;-\;\vec y\vec w^T}_F^2\;+\;\gamma J(w)\right)\;\;=\;\;0.
\end{eqnarray}
Due to $\textrm{Tr}[\vec A \vec A^T]=\normtwo{\vec A}_F^2$ , and the cyclicity of the trace ($\textrm{Tr}[\vec x \vec y^T] = \vec y^T \vec x$), the left-hand side can be written as

\begin{eqnarray*}
 \nabla_w \left(\textrm{Tr} \left[\vec X \vec X^T -  \vec X \vec w \vec y^T - \vec y \vec w^T \vec X^T + \vec y \vec w^T \vec w \vec y^T\right] \;+\; \gamma \vec w^T \boldsymbol{\Gamma} \vec w\right)
 &=& 0 - \nabla_w \left(2\textrm{Tr}\left[ \vec X \vec w \vec y^T\right]\;+\;\vec y^T \vec y \vec w^T \vec w + \gamma \vec w^T \boldsymbol{\Gamma} \vec w\right)\\
=\;\; \nabla_w \left( - 2\vec y^T \vec X \vec w \;+\; \vec y^T \vec y \vec w^T \vec w + \gamma \vec w^T \boldsymbol{\Gamma} \vec w  \right)\;&=&\;2\left(- \vec X^T \vec y\;+\;\vec y^T \vec y\vec w\;+\;\gamma \boldsymbol{\Gamma} \vec w\right).
\end{eqnarray*}

\noindent Solving (5) for $\vec w$ yields $\vec w = (\vec y^T \vec y \,\mathbb{I}\;+\; \gamma \boldsymbol{\Gamma})^{-1}\vec X^T \vec y$ , which is the row-vector equivalent to (2).

\end{document}